%% file: main.tex
\documentclass{article}

\usepackage{arxiv}
\usepackage[square,numbers,sort&compress]{natbib}
\usepackage[utf8]{inputenc} 
\usepackage[T1]{fontenc}    
\usepackage{hyperref}       
\usepackage{url}            
\usepackage{booktabs}       
\usepackage{amsfonts}       
\usepackage{nicefrac}       
\usepackage{microtype}      
\usepackage{lipsum}		
\usepackage{graphicx}
\usepackage{doi}
\usepackage{tabularx}
\usepackage{graphicx}
\usepackage{subcaption}
\usepackage{amsmath}
\usepackage{mdframed}
\usepackage{longtable}
\newcolumntype{P}[1]{>{\raggedright\arraybackslash}p{#1}}
\usepackage{comment}
\usepackage{csquotes}
\usepackage[table]{xcolor}
\usepackage{multirow}
\usepackage{array}
\usepackage{listings}
\usepackage{svg}
\usepackage{caption}
\usepackage{float}

\usepackage{orcidlink}
\usepackage{authblk} 

\usepackage{tikz}
\newcolumntype{C}[1]{>{\centering\arraybackslash}p{#1}}

\usepackage{soul}
\usepackage{xcolor}

\usepackage{fancyvrb}

\title{When Predictions Shape Reality: A Socio-Technical Synthesis of Performative Predictions in Machine Learning}

\author[1,2]{Gal Fybish\orcidlink{0009-0000-8931-7984}}
\author[1]{Teo Susnjak\orcidlink{0000-0001-9416-1435}}
\affil[1]{School of Mathematical and Computational Sciences, Massey University, Auckland, New Zealand}
\affil[2]{Corresponding author: Gal.Fybish.1@uni.massey.ac.nz}

\date{28/12/2025}
\renewcommand{\shorttitle}{When Predictions Shape Reality}

\begin{document}
\maketitle

\begin{abstract}
Machine learning models are increasingly used in high-stakes domains where their predictions can actively shape the environments in which they operate, a phenomenon known as performative prediction. This dynamic, in which the deployment of the model influences the very outcome it seeks to predict, can lead to unintended consequences, including feedback loops, performance issues, and significant societal risks. While the literature in the field has grown rapidly in recent years, a socio-technical synthesis that systemises the phenomenon concepts and provides practical guidance has been lacking.

This Systematisation of Knowledge (SoK) addresses this gap by providing a comprehensive review of the literature on performative predictions. We provide an overview of the primary mechanisms through which performativity manifests, present a typology of associated risks, and survey the proposed solutions offered in the literature. Our primary contribution is the ``Performative Strength vs. Impact Matrix" assessment framework. This practical tool is designed to help practitioners assess the potential influence and severity of performativity on their deployed predictive models and select the appropriate level of algorithmic or human intervention.

\end{abstract}


\input{Chapters/Introduction}




\input{Chapters/Background}


\input{Chapters/Methodology}

\vspace*{3cm}

\input{Chapters/Mechanisms-of-performativity}

\input{Chapters/Risks-Typology}

\input{Chapters/Solution-strategies}

\input{Chapters/PP-Extensions}

\input{Chapters/Performative-Matrix}

\input{Chapters/Discussion}

\input{Chapters/Conclusion}

\input{Chapters/List-of-Acronyms}

\bibliographystyle{unsrtnat}
\bibliography{references}  

\appendix

\clearpage

\end{document}

%% file: Chapters/Introduction.tex
\section{Introduction}

Predictive models are rarely isolated from their operating environment. When deployed, their outputs inform decisions that can reshape the very outcomes they aim to predict. Banking is a prime example of this phenomenon. A bank's lending scoring model may predict that an applicant is at high risk of default, and, as a result, the bank assigns a high interest rate to the loan. This higher rate increases the financial burden on the applicant, which in turn can cause the very default the model predicted - a classic self-fulfilling prophecy. 
This is an example of what Perdomo et al. coined “Performative Predictions”: a phenomenon in which model-driven decisions alter the data-generating process in a way that future observations depend on the model itself \citep{perdomo_performative_2020}. In this context, data is not a static reflection of the world but is actively influenced by the predictions we publish \citep{perdomo_revisiting_2025}. The consequences of this dynamic can be significant, including performance degradation, the entrenchment of systemic biases, and an erosion of trust in predictive systems.

Several summary studies have been published covering various aspects of the field of performative predictions. A taxonomy of bias in data and its relation to the performativity of predictive models was proposed by \citet{pombal_prisoners_2022}, while \citet{pagan_classification_2023} presented a comprehensive definition and taxonomy of feedback loops and their relation to bias. A recent work by \citet{khosrowi2025predictions} provided an overview of the field, highlighting ethical challenges and calling for a coordinated research effort to address issues arising from performative predictive models. However, to the best of our knowledge, there is no published socio-technical synthesis of knowledge in the field that: (i) systematises mechanisms and risks, (ii) organises solution strategies across algorithmic and governance layers, and (iii) offers a methodology to reason about real-world use cases. Specifically, practitioners lack a straightforward method to assess the nature and the severity of performativity in a given use case and select an appropriate strategy to manage it. This work aims to close this gap. Thus, 
our contribution is threefold:
\begin{enumerate}
  \item We present a comprehensive explanation of the mechanism of performative predictions and the risks associated with them
  \item We survey solution strategies published in the academic literature of the field.
  \item We introduce the \textbf{Performative Strength vs. Impact Matrix}. This practical framework is designed to help practitioners understand the influence of deploying predictive models and enable them to make informed decisions on how to manage performativity.
\end{enumerate}

To make the abstract concepts in the work concrete, we will use two running examples from the high-stakes domain of clinical prediction. The first example is the \textbf{hospital readmission models} \citep{feng_bayesian_2022}. Hospitals utilise predictive models to estimate the likelihood of a patient returning to the hospital within a short period after discharge. A high-risk prediction from such a model may be used to trigger a preventative intervention, which aims to prevent the readmission \citep{feng_designing_2024}. This use case illustrates a dynamic where the model's prediction is negated by the action it inspires. Our second, contrasting example, is the \textbf{prognostic mortality model}, often used to assess the futility of care or the likelihood of death \citep{adam_addressing_2024} \citep{amsterdam_when_2025}. A prediction of a high chance of mortality can drive a clinical decision to withdraw life-saving treatments and shift to supportive care, a decision that in turn can cause mortality \citep{adam_addressing_2024}. This illustrates the opposite: a self-fulfilling dynamic in which the model's prediction causes the very outcome it forecasts. 

The remainder of the paper is organised as follows: Section 2 provides background on the core concepts of performative predictions. Section 3 outlines the methodology for our review, including the research questions and the process for selecting papers. In Section 4, we describe the primary mechanisms through which performativity manifests. Section 5 presents a typology of the risks associated with performativity. Following this analysis, Section 6 surveys the landscape of proposed solutions found in the literature. Section 7 presents extensions to the core context of performative predictions. In Section 8, we introduce our novel contribution: the Performative Strength vs. Impact Matrix, a framework for assessing real-world use cases. Finally, Section 9 discusses the implications of our work and outlines future research directions, and Section 10 concludes the paper.

%% file: Chapters/Background.tex
\section{Background}

While the concept of performativity has long been explored in fields such as economics and linguistics, it remains relatively novel in the context of predictive models \citep{perdomo_performative_2023}. In supervised machine learning, performativity can lead to distribution shifts and is primarily addresses through model retraining \citep{perdomo_performative_2020}.

Aside from ones already mentioned, other examples include predictions of stock prices that can influence trading decisions and affect stock prices \citep{perdomo_performative_2020}, as well as forecasts regarding climate that can inform policies that may impact the environment in the future \citep{perdomo_revisiting_2025}.
When considering the connection between predictions and the environments in which they operate, it becomes evident that performative predictions are common and occur whenever a model’s prediction concerns people.
Accepting the performative nature of these models can lead to more accurate forecasting and finding ways to channel them for more favourable social outcomes \citep{perdomo_performative_2023, mishler_fair_2022}. In certain situations, the goal of a prediction is to influence its environment; for instance, when predicting the probability of a person having a medical condition, with the assistance of a timely prediction, we aim to prevent it \citep{perdomo_performative_2023}.

\begin{figure}[htbp]
    \centering
    \includegraphics[width=0.6\textwidth]{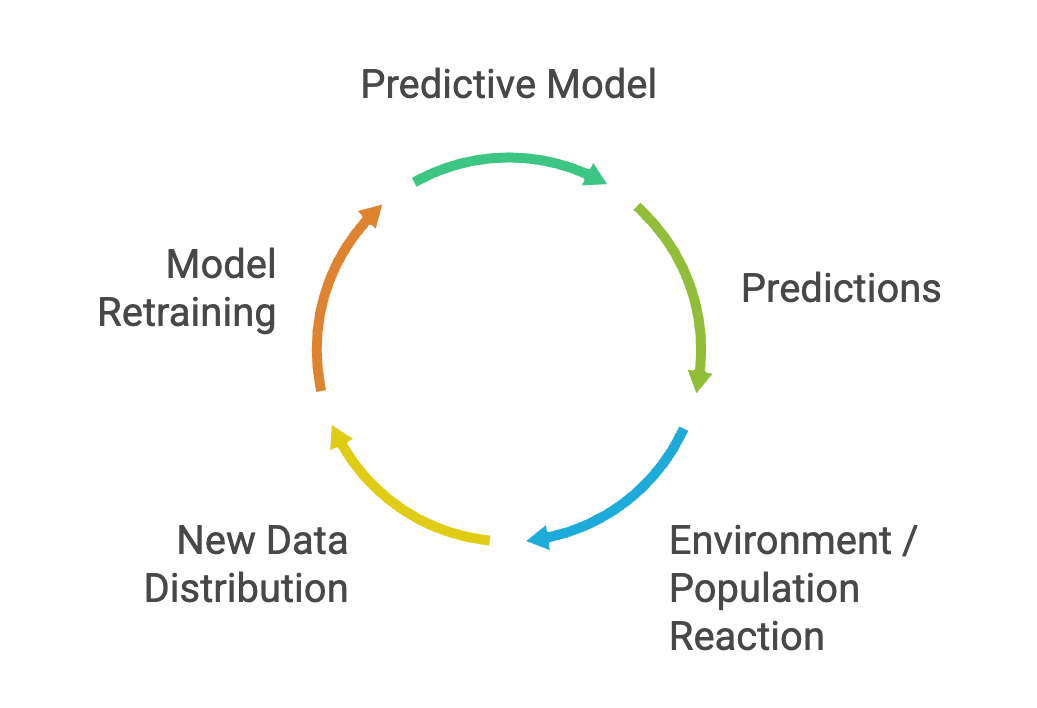}
    \caption{The Performative Prediction Cycle. A deployed model's predictions influence the environment, which in turn generates a new data distribution used for future model retraining.}
    \label{fig:perf_cycle}
\end{figure}

Supervised machine learning models, which are widely used for predictions, assume that their data distribution is fixed; therefore, the predictions made by these models cannot alter that distribution. However, actions based on these predictions can influence their environment, thereby contradicting this assumption and potentially degrading the models' performance \citep{mofakhami_performative_2023}. To formally address the challenge of performativity in machine learning models, the field introduced several key concepts:

\textbf{Performative Prediction} represents the notion that machine learning models' predictions do not just passively forecast an outcome, but actively influence or cause that outcome. The very act of making a prediction alters the environment in which the model operates and, in turn, changes the data that subsequent iterations of the model will encounter in the future \citep{perdomo_performative_2020}. This dynamic is illustrated in Figure \ref{fig:perf_cycle}.

\textbf{Distribution map $D(\theta)$} is a function that uses the parameters of a predictive model $(\theta)$ and maps them to a new data distribution that emerges after the model has been deployed and its predictions influence the environment \citep{peet-pare_long_2022}.

\textbf{Performative Risk} (PR) is a measure of a model's performance that explains the fact that the model's predictions and actions taken based on these predictions can change the data distribution from which future data will be drawn. 

In a non-performative context, it is assumed that the model relies on a fixed underlying data distribution. Conversely, in the performative context, this assumption no longer holds as the model itself induces a change in the data distribution. In this case, the Performative Risk of a predictive model with parameters $(\theta)$ is defined as follows:
\[ 
PR(\theta) = \mathbb{E}_{z \sim D(\theta)} \left[ \ell(z; \theta) \right] 
\]
i.e, the performative risk is the expected loss of a predictive model with parameters $(\theta)$, calculated over the data distribution $D(\theta)$ that has been induced by deploying the predictive model.

Performative Risk was introduced by \citet{perdomo_performative_2020} to describe the loss function of a predictive model relative to the data distribution created as a result of its deployment. This differs from traditional modelling, which assumes a fixed distribution over its input features and target variable \citep{perdomo_performative_2020}.
Traditionally, the model risk minimisation aims to find a set of model parameters that minimises the loss function over the fixed distribution \citep{perdomo_performative_2023}; however, under conditions of performativity, the aim is to minimise the loss over the distribution created as a result of the model's deployment and not the model's original distribution \citep{peet-pare_long_2022}.

The recognition of performativity introduces a tension in modelling objectives. A distinction exists between the need for accuracy and the desire to influence the environment toward a specific outcome \citep{kim_making_2023}. This has led to different perspectives on how to manage performativity. Two opposing approaches have been offered by \citet{khosrowi_managing_2023}: an appraisal view and a mitigation view. The appraisal view sees performativity as potentially positive.
In contrast, the mitigation view calls for counteracting the effects of performativity by modelling the responses to the predictions and adjusting the model accordingly. According to \citet{khosrowi_managing_2023}, neither approach is satisfactory. The appraisal view may allow values to shape models in ways that undermine their credibility, whereas the mitigation view may deny the potential benefits of performativity.

Related concepts to performative prediction have also been introduced. ``Performative Power'', introduced by \citet{hardt_performative_2022} as a measure of the impact firms can have on people's behaviour. Using predictive models, firms with high performative power can steer populations toward outcomes that are more profitable for them.
“Outcome Performativity” has been used by \citet{kim_making_2023} to describe instances in which focused decisions affect specific outcomes for individuals, rather than the effects of more general decisions on the population’s data distribution.

%% file: Chapters/Methodology.tex
\section{Methodology}

\subsection{Research Questions}

We structure our review around three research questions. These questions are not independent; they are designed to follow a logical \textbf{Cause $\rightarrow$ Effect $\rightarrow$ Response} progression that forms the narrative backbone of this SoK. This structure allows us to map the field of performative predictions systematically:

RQ1 - What are the mechanisms through which performative predictions manifest?

RQ2 - What are the risks associated with performative predictions?

RQ3 - What strategies are used to mitigate the risks associated with performative predictions?

\subsection{Papers Selection}

To identify relevant studies for this SoK, we searched Discover, Scopus, and Google Scholar in May 2025. After experimenting with several search strings, we used the search string ``Performative AND prediction*" for the Discover and Scopus databases. Using Google Scholar, we used the search string  ""performative prediction" AND "machine learning"". For all searches, we limited the results to English-language publications published between 2019 and 2025.

Inclusion: we included works that (i) explicitly discuss performative predictions or closely related notions in machine learning; (ii) present formal analysis, empirical evaluation, or conceptual frameworks related to RQ1-RQ3; and (iii) are peer-reviewed conference/journal papers or credible preprints, and published thesis works.

Exclusion: we excluded non-scholarly works, items lacking sufficient bibliographic details, non-English works, studies whose focus is unrelated to performativity in machine learning, and early versions of published papers.

\begin{figure}[h!]
  \centering
  \includesvg[width=0.5\linewidth]{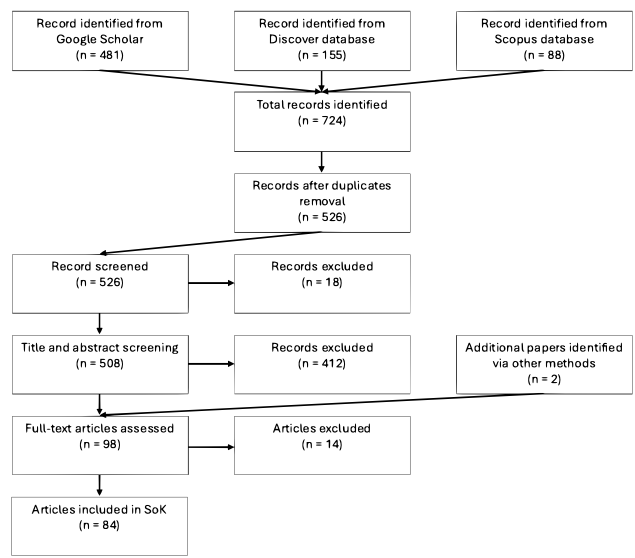}
  \caption{Literature search flow diagram}
  \label{fig:literature-search-strategy}
\end{figure}

The database queries returned 724 results, which were then checked for duplications, both within each source and between sources. After removing the duplicate results, we were left with 526 records for initial screening. During the initial screening process, we excluded 18 records due to missing information, being written in a language other than English, or not being a published paper. After the initial screening, the titles and abstracts of 508 published works were screened for their relevance to the SoK, after which 412 works were excluded. The remaining 96 works, along with two other works identified through different methods, underwent full-text assessment. During this assessment, 14 works were excluded for irrelevance to the SoK or poor quality. After this process was completed, we decided to include two additional papers published after the database search, bringing the number of published works included in the SoK to 84. The selection process is presented in Figure~\ref{fig:literature-search-strategy}.

To reduce the risks of missed coverage and topic drift, we performed limited backward/forward snowballing from several key papers identified during the investigation. This added two works, which are captured under the "identified via other methods" count.

%% file: Chapters/Mechanisms-of-performativity.tex
\section{Mechanisms of performativity}

This section surveys the primary mechanisms through which performativity manifests. These mechanisms, the feedback loop and data shifts, are the root causes of the performance and societal risks detailed in Section 5.

\subsection{Feedback Loops}

When the model's predictions affect its environment, thereby changing the input for future training cycles, a feedback loop is created. In this dynamic, the model and the environment become increasingly reliant on each other \citep{makowski_feature_2024}, which can increase the risk of bias in the model's results \citep{pagan_classification_2023}.
Feedback loops are common in many use cases, such as product recommendations and medical diagnostics, but are less prevalent in others, such as weather prediction \citep{khritankov_positive_2023}.

In their comprehensive work,  \citet{pagan_classification_2023} employed the dynamic systems methodology to represent the process of utilising machine learning models as feedback loops. The authors differentiated between open and closed feedback loops and established a formal classification of feedback loops based on their impact on the machine learning process.
In their work, \citet{pagan_classification_2023} distinguished between several types of feedback loops, which can occur in combination with each other in the same machine learning problem space:

\textbf{Sampling Feedback Loop} - 
Decisions made on the basis of data sampled from different populations can create a feedback loop, whereby the relative size of the populations changes over subsequent iterations of the model's training, leading to decreased sampling from specific populations and, in extreme cases, their complete disappearance from the training data.

\textbf{Individual Feedback Loop} - 
In this type of feedback, the decision affects an individual's characteristics, which are then used in the subsequent training of the model.

\textbf{Feature Feedback Loop} - 
In a feature feedback loop, a decision made as a result of a model's output affects the value of a feature used in the training dataset, leading to a new decision that, in turn, affects the value of the feature.

\textbf{ML Model Feedback Loop} - 
A model feedback loop occurs when the training or validation data of the model depends on decisions made based on prior predictions of the model. For instance, lending decisions made using predictive models can lead to future data that includes only cases where loans were approved. 

\textbf{Outcome Feedback Loop} - 
An outcome feedback loop occurs when the decision affects the outcome, which is then fed back to future training of the model. For instance, a decision to approve a loan, albeit at a higher interest rate, can increase the probability of default, thereby affecting the target feature of the model. 

\textbf{Adversarial Feedback Loops} - 
This type of feedback loop occurs when individuals can react to the decisions made as a result of a predictive model and influence the feedback process \citep{drusvyatskiy_stochastic_2023,he_decision-dependent_2025}, and is also widely known as \textbf{Strategic Classification}. The result is an interplay between the deployed model and the population affected by it, which reacts in ways that changes the model's predictions in their favour \citep{perdomo_performative_2023}. For instance, in our hospital readmission model, a patient who knows that ``high-risk" individuals receive additional follow-up care might exaggerate their symptoms or lack of social support, thereby altering their "features" during their discharge interview. The goal of the patient is to ensure classification as ``high-risk" to receive their desired preventive intervention. 
In most situations, the reacting population is assumed to react in a rational way that will be most beneficial to them \citep{hardt_performative_2022}. The performative effect of strategic classification is primarily centred on the predictive model's features, while the model's probability distribution is usually assumed to be unchanged \citep{mendler-dunner_anticipating_2022}. 

Feedback loops can exhibit different dynamics. They may be a \textbf{Self-fulfilling feedback loop}, where the model induces decisions that confirm its own predictions, leading to more instances of the predicted outcome over time \citep{gower-winter_identifying_2025}. Our prognostic mortality model is a clear example: a prediction of a high probability of mortality can lead a clinical team to provide the patient with supportive treatment instead of life-saving care \citep{adam_addressing_2024}, which, in turn, causes the patient to die, thereby fulfilling the prophecy. 
In contrast, feedback loops can be \textbf{Self-negating}, where actors can react in a way that prevents the predicted outcome, resulting in fewer instances of the expected outcome in future model iterations \citep{gower-winter_identifying_2025}. Our hospital readmission model example illustrates this: a ``high-risk" prediction triggers a preventive follow-up intervention that prevents the readmission, thereby negating the prediction.

In a study connecting feedback loops to concept drifts,  \citet{khritankov_positive_2023} argued that \textbf{Positive feedback loops} are created when a model's predictions are used as inputs for subsequent predictions, and over time cause a concept drift. The feedback loop may occur when the training data is procured from the same population that later relies on the model's predictions, or in cases where the environment is affected by the users' behaviour \citep{khritankov_positive_2023}. The effects of the feedback loop may not be instantly apparent and only become observable after the model has been deployed and utilised for some time \citep{khritankov_positive_2023}.
The primary consequence of these feedback loops is that they cause the statistical properties of the data to change, leading to the shifts in data and distribution discussed next.

\subsection{Data Shifts}

The conventional assumption in machine learning is that the data distribution is static and fixed throughout the model life cycle \citep{izzo_theory_2023}. However, the deployment of predictive models in real-world scenarios often violates this core premise, leading to an evolving data distribution that can cause deterioration in the model performance \citep{izzo_how_2021,chislett_ethical_2024}. This phenomenon is broadly called \textbf{Concept Drift} \citep{gower-winter_identifying_2025} or \textbf{Distribution Drift} \citep{izzo_theory_2023,wyllie_fairness_2024}, and is defined as any change in the underlying data-generating process over time \citep{gower-winter_identifying_2025}.

Such changes in the data distribution can originate from two distinct types of sources: \textbf{External}, or \textbf{Exogenous}, changes in the input data that are caused by factors outside of the deployed model, and to a large extent, are independent of it, such as environmental or temporal changes \citep{shan_beyond_2023}. In contrast, \textbf{Internal}, or \textbf{Endogenous}, data shifts are caused by decisions or actions resulting from the deployment of a predictive model \citep{he_decision-dependent_2025,yan_decentralized_2024}.

Performativity is the core mechanism of endogenous data shifts, characterising how the predictive model actively influences the data distribution it aims to forecast \citep{peet-pare_long_2022}. This model-induced change is specifically referred to as \textbf{Performative Drift (PD)}, which is recognised as a subtype of Concept Drift \citep{gower-winter_identifying_2025}. Performative Drift manifests through two distinct mechanisms: \textbf{Concept Shift} involved a change in the underlying relationships between the model's features and outcomes \citep{mishler_fair_2022,makowski_feature_2024}; conversely, \textbf{Covariate Shift} entails a change only in the distribution of the features, while the relationship between the features and the outcomes stay the same \citep{mishler_fair_2022}. 

To illustrate these concepts, we can use our running examples from the clinical domain. In this domain, population ageing constitutes an \textbf{Exogenous} shift, as it changes the data distribution but is not caused by a model's deployment. A predictive model's deployment, however, may cause \textbf{Endogenous} shifts. For example, a predictive patient triage system may alter the arrival patterns of patients, creating a \textbf{Covariate Shift}; here, the distribution of incoming patients changes, but the medical relationship between their symptoms and conditions remains stable. In contrast, our readmission model example illustrates \textbf{Concept Shift}. Here, a high-risk prediction may trigger a preventative intervention. If this intervention is successful and prevents readmission, then it changes the original relationship between the patient's features and the outcome.

%% file: Chapters/Risks-Typology.tex
\section{Performative Predictions Risks Typology}

\begin{figure}[h!]
    \centering
    \includegraphics[width=0.6\textwidth]{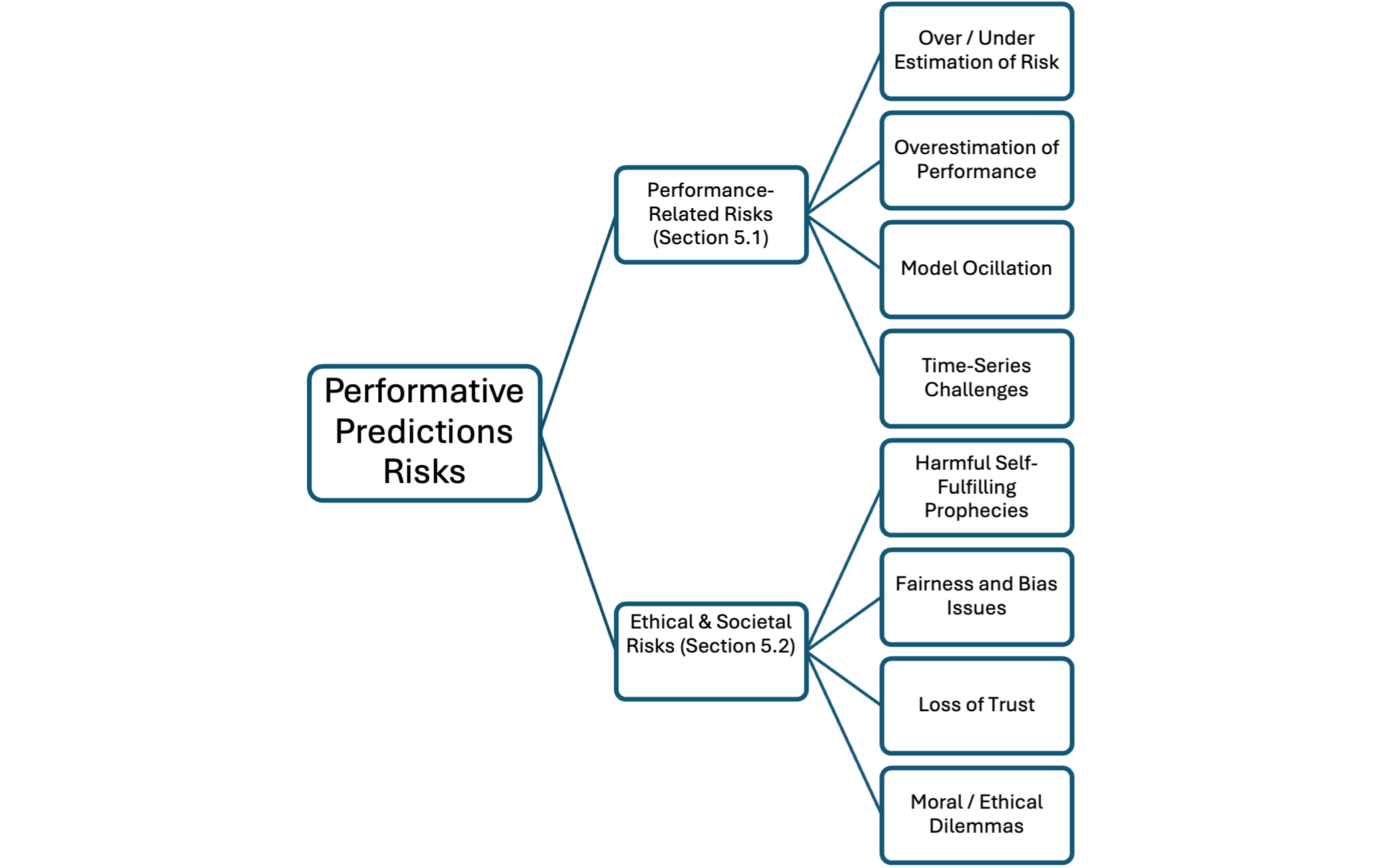}
    \caption{Performative Predictions Risks Typology}
    \label{fig:risks_typology}
\end{figure}

The mechanisms of performativity described in the previous section, comprising feedback loops and their resulting data shifts, are the direct cause of a spectrum of risks when predictive models are deployed. These risks are not merely theoretical; they can degrade model performance, mislead practitioners, and create significant societal harm. 
This section presents a typology of these risks, which we separate into two closely related categories. We first discuss the \textbf{performance-related risk} (Section 5.1): the immediate, technical failures, such as statistical misestimation, inaccurate metrics, and instability. We then examine the broader \textbf{ethical and societal risks} (Section 5.2): the human-centric, real-world harms, such as bias entrenchment, harmful prophecies, and loss of trust, often caused or amplified by the underlying technical failures. Figure \ref{fig:risks_typology} provides a visual map of this taxonomy, which we discuss in detail below.

\subsection{Performance-related Risks}

This category covers the technical failures and instabilities arising from performativity.

\textbf{Over or under estimation of risk} - Performative feedback loops can cause a model to develop a skewed view of the data-generating process, leading to incorrect estimation of risk.

\begin{itemize}
    \item \textbf{Overestimation}: This often happens in self-fulfilling loops. A feedback loop, caused by the model's performativity, can lead to the model exhibiting higher-than-acceptable rates of false-positive predictions \citep{adam_addressing_2024}. For instance, in our prognostic mortality model example, if the model incorrectly predicts that a patient will die, the patient might be de-prioritised for a life-saving treatment and subsequently die. Retraining the model on this outcome will reinforce its incorrect prediction, leading it to overestimate the true risk for similar patients \citep{adam_addressing_2024}.
    \item \textbf{Underestimation}: This happens in self-negating loops. When prediction-based interventions are successful, as can potentially happen in our readmission model example, retraining the model using this ``good" outcome can cause the model to underestimate the true, underlying risk for future patients who may not receive the intervention \citep{boeken_evaluating_2024}\citep{chislett_ethical_2024}.
\end{itemize}

\textbf{Overestimation of performance} - 
When a model's prediction affects the data distribution, its performance metrics can become misleadingly inflated. Practitioners who rely on these metrics may believe the model is performing much better than it actually is, leading them to act on its predictions with false confidence \citep{adam_addressing_2024}.

\textbf{Model oscillation} - 
A performative model may oscillate, meaning it will change the predicted class after continuous retraining and cause deterioration in the model's predictive capability and stability \citep{adam_addressing_2024}.

\textbf{Challenges in time-series forecasting} - 
The effects of performative predictions are particularly apparent in time-series models, which use past observations to predict future observations \citep{bhati_performative_2022}. When actions are taken based on these predictions, future observations are influenced by the actions, thereby partially obscuring the actual data distribution of the modelled phenomenon \citep{bhati_performative_2022}. In addition, under conditions of performativity, the distribution of some features that are part of the time-series model can change as a result of the model's deployment, thus increasing the challenge of accurate forecasting \citep{zhao_performative_2023}.

\subsection{Ethical and societal risks}
 The use of predictive models can give rise to several ethical and societal risks, including fairness, bias, trust, and moral dilemmas. Before the performative prediction discourse developed, work on model bias and fairness tended to focus on static data environments \cite{somerstep_algorithmic_2024}. The dynamic environments in which many predictive models operate necessitate a shift in how bias and fairness are considered. Following, we cover the human-centric consequences of performative models. 

\textbf{Harmful self-fulfilling prophecies} - 
In certain situations, decisions made based on a predictive model can lead to unintended harm through self-fulfilling prophecies \citep{amsterdam_when_2025}. For instance, prioritising aggressive cancer treatments for patients with slow-growing tumours over those with fast-growing tumours, based on a predictive survival model, can result in reduced survival chances for patients with the fast-growing tumours \citep{amsterdam_when_2025}. This phenomenon has been empirically observed in the medical literature, where this kind of ``prophecies" in resuscitation decisions have been shown to directly influence patients' survival rates \citep{de-arteaga_self-fulfilling_2023}.

\textbf{Fairness and bias issues} - 
The use of predictive models needs to ensure fairness and adequate representation of diverse populations \citep{jin_addressing_2024}; however, in practice, the performative nature of predictive models can cause unfairness when they are used, for instance, in policing or college acceptance decisions \citep{peet-pare_long_2022,pagan_classification_2023}. The issue of fairness and bias could have severe implications in the example of the prognostic mortality model. If the model's training data reflects historical biases, e.g. that marginalised groups received less-aggressive care, it may learn to associate those groups with futility. The resulting self-fulfilling loop will entrench the bias against marginalised groups \citep{amsterdam_when_2025}.

The deployment and use of predictive models can introduce bias in the data due to the model reshaping the data distributions. For instance, using a predictive model for fraud detection can lead to genuine requests for opening a bank account being rejected, potentially causing unfairness and bias in the data used for future predictions \citep{pombal_prisoners_2022}. Changes to the data distribution can cause models that were adequately trained to avoid bias and unfairness to become biased after being deployed, even with the introduction of policies intended to address bias \citep{mishler_fair_2022,pagan_classification_2023}. Without proper mitigation, predictive models can exhibit lower prediction accuracy for minority groups compared to majority groups \citep{peet-pare_long_2022}.
 
The standard solutions to performative predictions can cause severe representation and fairness issues due to certain groups overshadowing the minority groups \citep{jin_addressing_2024}. When models are trained using the results of previously deployed models, they tend to converge and rely on the majority population more and more, resulting in under representation of minority groups \citep{wyllie_fairness_2024}.

Furthermore, the long-term societal impact of these feedback loops can be profound. As studied by \citet{lankireddy_when_2024}, online predictive systems can, under certain conditions, lead to outcomes such as preference polarisation or consensus within the affected population, demonstrating how model dynamics can shape collective behaviour over time.
 
\textbf{Adoption and loss of trust risk} - 
When a predictive model is designed to support a decision-making process, trust in the model's capabilities plays an important role in the usability and acceptance of the model \citep{adam_addressing_2024}. In a performative environment, potentially accurate predictions made by the model may not materialise, potentially eroding trust in future predictions. This can be the case in pandemic predictions, where actions taken to limit the spread of a virus may successfully reduce the pandemic's effect on the population. Because the predicted outcome did not materialise, the public may believe the model was wrong, and have less trust in future predictions \citep{gois_performative_2024}.

\textbf{Moral and ethical dilemmas for modellers and policy makers} - 
The use of performative models and the solutions designed to manage them raises moral and ethical dilemmas. For example, in the readmission model, hold-out data sets can be created from subsets of patients who are chosen not to receive a model-recommended treatment for data-gathering purposes. However, this practice raises several potential ethical issues that need to be considered before it \citep{chislett_ethical_2024}.

Furthermore, the performative attribute of predictive models can be used to steer outcomes; however, this raises ethical questions about where the model needs to steer \citep{khosrowi2025predictions}. In these situations, decision-makers may need to choose between forecasting accuracy and steering towards improved outcomes \citep{kim_making_2023}. As performativity, through informing decisions, has the power to change outcomes, it is paramount to find models that not just predict outcomes accurately but also steer them towards socially desired outcomes \citep{perdomo_performative_2023}.

%% file: Chapters/Solution-strategies.tex
\section{Solution Strategies}
\begin{figure}[h!]
    \centering
    \includegraphics[width=0.6\textwidth]{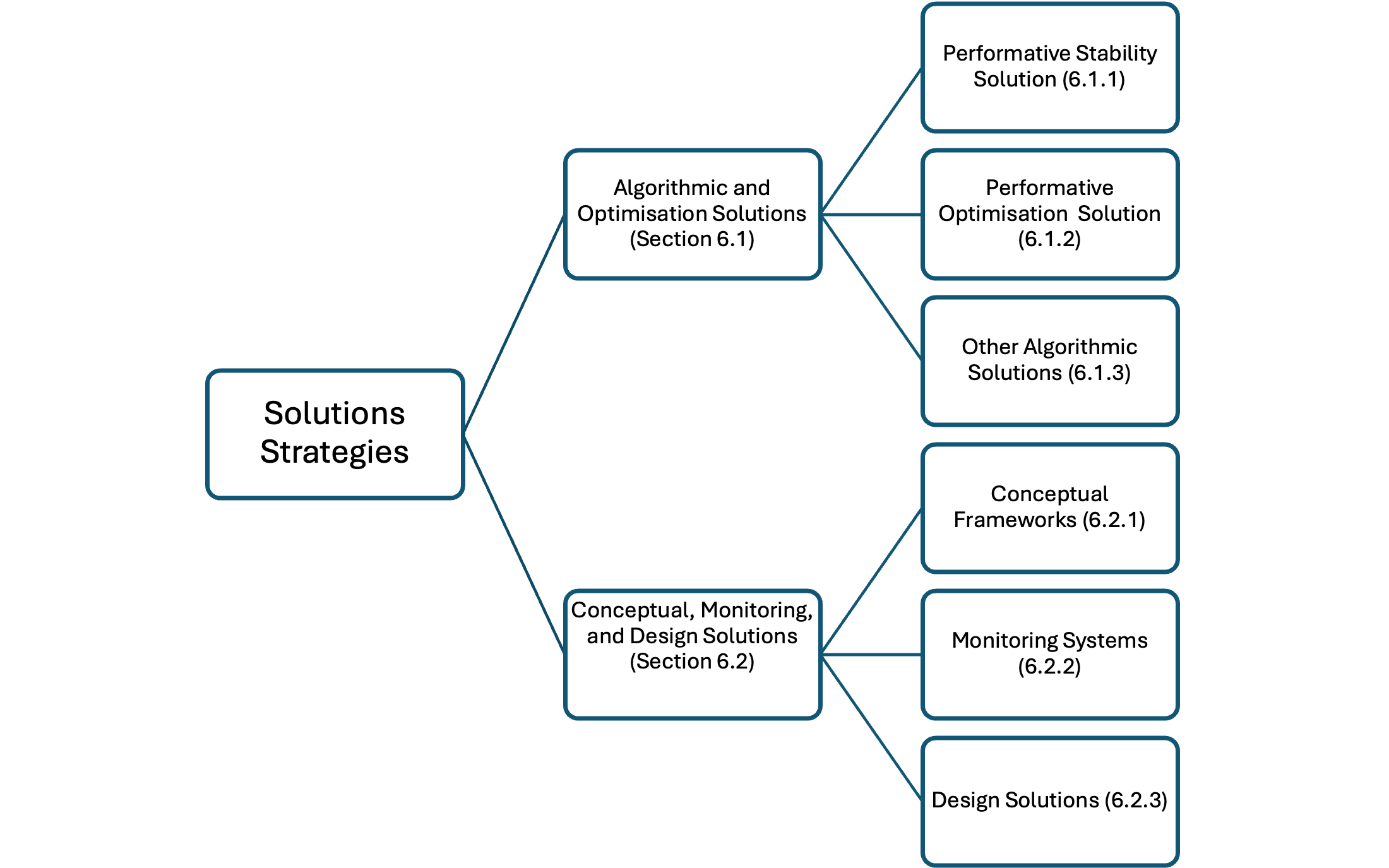}
    \caption{Solution Strategies Tree}
    \label{fig:solutions_tree}
\end{figure}

The challenges posed by the mechanisms (Section 4) and risks (Section 5) of performativity have led to the development of various solution strategies. This section provides a comprehensive overview of these solutions, which are designed to address and mitigate performative effects.  
The proposed solutions span a wide range, from formal algorithmic methods to broader conceptual and systemic interventions.
To provide a comprehensive overview, this section categorises these approaches into two primary branches illustrated in Figure \ref{fig:solutions_tree}. First, Section 6.1 details the ``Algorithmic and Optimisation Solutions" that address performative risk mathematically. Second, Section 6.2 surveys the complementary ``Conceptual Re-Framing, Monitoring, and Design Solutions", which cover the non-algorithmic approaches for managing performativity in practice.

\subsection{Algorithmic And Optimisation Solutions}

As previously discussed, Performative Risk refers to the loss function of a predictive model in relation to the data distribution that results from its deployment. 
To solve the issue of Performative Risk, \citep{perdomo_performative_2020} introduced two new concepts, \textbf{Performative Stability} and \textbf{Performative Optimality}.
Performative stability aims to find a model that is optimal over the distribution it created, where there is no need for further retraining \citep{gower-winter_identifying_2025}. If we retrain the performative stable model again with its induced distribution, it will return the same model \citep{perdomo_performative_2020}.
A performative-optimal model aims to minimise the model's performative risk \citep{perdomo_revisiting_2025}. In this case, the model minimises the performative risk across all the models that can be used over the data distribution \citep{khosrowi2025predictions}.

A performative optimal model for any distribution does not need to be the same as a performative stable model for the same distribution. A performative optimal model is not necessarily performative stable, and a performative stable model is not necessarily performative optimal \cite{perdomo_revisiting_2025}. 

In order to systematise the knowledge, we first survey solutions for achieving Performative Stability (Section 6.1.1), then examine methods for finding a Performative Optimal point (Section 6.1.2), and conclude with other related algorithmic approaches (Section 6.1.3). These diverse approaches are summarised and compared in Table \ref{tab:algo_solutions} at the end of this section. 

\subsubsection{Performative Stability Solutions}

The first algorithmic goal in managing performative predictions is to find a \textbf{Performative Stable} point. In this equilibrium, training the model on the data distribution it induces yields the same model \citep{perdomo_performative_2020}. Achieving stability ensures that the model's behaviour settles down despite its influence on the environment.

\textbf{Repeated Risk Minimisation} - 

The foundational approach to achieving stability is \textbf{Repeated Risk Minimisation (RRM)}, first proposed by \citep{perdomo_performative_2020}. 
The RRM procedure iteratively retrains the predictive model using data drawn from the distribution created by deploying the previous iteration of the model \citep{hardt_performative_2023}. If this iterative process converges, the resulting model is performatively stable, as further retraining would not change it \citep{perdomo_performative_2020}. RRM has been shown to achieve performative stability not only in supervised learning contexts but also for neural network models \citep{mofakhami_performative_2023}.
 
However, simple RRM has limitations. It may fail to converge, or converge to a stable point that is far from optimal \citep{kabra_limitations_2024}. To address these issues, several extensions have been proposed. \textbf{Regularised Repeated Risk Minimisation (Reg-RRM)} introduces regularisation to slow the retraining pace and prevent large jumps between iterations \citep{kabra_limitations_2024}. When the underlying distribution is uncertain, \textbf{Repeated Robust Risk Minimisation (R\textsuperscript{3}M)} uses a set of potential distributions that are centred around a reference distribution to the real, unknown one, to achieve faster convergence \citep{jia_distributionally_2024}. \textbf{Affine Risk Minimising} enhances convergence by incorporating results from multiple previous training steps, rather than just the immediate predecessor \citep{khorsandi_tight_2024}. RRM has also been extended to \textbf{bi-level} machine learning problems, where the input distribution of each level depends on the outputs of the other level \citep{lu_bilevel_2023}. In this class of problems, it is necessary to solve two risk-loss minimisation problems, thereby requiring the attainment of Bi-Level Performative Stability (BPS). This can be done using the \textbf{Bi-level Repeated Risk Minimisation (Bi-RRM)} procedure, or more efficiently using \textbf{Bi-level Stochastic Gradient Descent (Bi-SGD)} \citep{lu_bilevel_2023}.

Despite these improvements, RRM-based approaches continue to face criticism. They may implicitly favour models with less data variability, potentially leading to increased bias or convergence towards outlier values \citep{tsoy_impact_2025}. Furthermore, RRM could converge to different stable points depending on initial conditions \citep{dong_approximate_2023}. To counter potential unfairness, \textbf{Repeated Distributed Robust Optimisation (RDRO)} combines RRM with Distributed Robust Optimisation (DRO) to ensure stability while protecting underrepresented groups \citep{peet-pare_long_2022,peet-pare_beyond_2022}.

\textbf{Stochastic Gradient Descent} - 

As an alternative to the potentially costly full retraining required by RRM, stochastic gradient descent (SGD) methods update the model's parameters using a single gradient step on the loss function \citep{perdomo_performative_2023}. Introduced by \citet{mendler-dunner_stochastic_2020}, SGD approaches can be \textbf{greedy} (i.e., updating after every new data point) or \textbf{lazy} (i.e., updating after several new data points). Both are shown to converge to performative stability, with the choice between them depending on the strength of performativity and deployment costs \citep{mendler-dunner_stochastic_2020}. \textbf{SGD with greedy deployment (SGD-DG)} was also studied by \citet{li_stochastic_2024}, who demonstrated convergence to stable solutions for non-convex loss functions. This line of research has been further developed by \citet{drusvyatskiy_stochastic_2023}, who showed how stochastic methods, initially developed for non-performative situations, can be applied to performative models and converge to a performative stable point. Further refinements include \textbf{Clipped-SGD} \citep{li_clipped_2024} and analysing the process through \textbf{Stochastic Forward-Backward(SFB)} dynamics \citep{cutler_stochastic_2024}. SGD has also been applied to state-dependent performativity, where agents react to the deployment of a predictive model based on previous states rather than just the latest one \citep{li_state_2022}. Following this work, \citet{brown_performative_2022} used RRM and a delayed (lazy) version of it to converge to a stable solution in a state-dependent performative situation. 

Finally, the concept of performative stability has also been extended to multi-agent settings, where multiple models may compete or cooperate. In these scenarios, stability is often defined as a Nash Equilibrium, a state in which no agent can improve its outcome by unilaterally changing its strategy. Stability can be achieved through methods similar to the single-agent case, such as repeated training \citep{narang_learning_2022}, or by using derivative-free or adaptive stochastic methods to find the equilibrium \citep{narang_multiplayer_2023}. Cooperative multi-agent systems can use a decentralised extension to SGD (\textbf{DSGD-GD}) to find a joint stable point \citep{li_multi-agent_2022}, and network effects where agents learn from each other's deployment have also been modelled \citep{wang_network_2023}.

While the stability-seeking methods discussed above offer a path to equilibrium, they face a critical limitation: a stable model is not necessarily optimal. An algorithm could converge to a performative stable point that is highly sub-optimal, or even one that maximises the performative risk \citep{miller_outside_2021}. This fundamental issue — that stability does not equal optimality — motivated seeking \textbf{Performative Optimality}, which is the focus of the next section.

\subsubsection{Performative Optimisation Solutions}

Recognising that stability is insufficient, a significant body of work seeks \textbf{Performative Optimality (PO)} - finding the model's parameters that minimise the performative risk. One of the main challenges here is that the actual distribution map induced by the predictive model's deployment is unknown \citep{miller_outside_2021}.

The first approach to PO, presented by \citep{miller_outside_2021}, involved a two-stage process: first estimating the distribution map, then optimising a surrogate to the performative risk, considering the estimated distribution map as the true one. Building on these \textbf{Performative Gradient Descent (PerfGD)} was developed by \citep{izzo_how_2021} to directly optimise the performative risk by estimating its gradient, often outperforming stability-seeking methods. Extensions include \textbf{Stateful Performative Gradient Descent (Stateful PerfGD)} for environments where the distribution changes gradually rather than instantly \citep{izzo_how_2022,izzo_theory_2023}. Further advances, such as the push-forward model, which was accompanied by a novel estimator for the performative risk gradient, the \textbf{Reparametrisation-based Performative Gradient (RPPerfGD)}, allow for better estimation of the performative risk function gradient, facilitating more efficient and scalable methods to find the performative optimal point \citep{cyffers_optimal_2024}. These gradient-based methods have been shown to work even under relaxed convexity assumptions \citep{zhao_optimizing_2022}.
  
Other strategies address the challenge of unknown distribution maps in different ways. \textbf{Distributionally Robust Optimum (DRPO)} extends DRO concepts to efficiently handle cases where the assumed distribution map differs from the true underlying one \citep{xue_distributionally_2024}. For environments with delayed, geometrically decaying dynamics, an iterative approach is used to deploy the model multiple times, allowing the evolving distribution to stabilise before applying a gradient update \citep{ray_decision-dependent_2022}.
 
Alternatively, the optimisation problem can be framed using an online learning approach for regret minimisation \citep{jagadeesan_regret_2022}, with practical implementations using parameter-free models \citep{park_parameter-free_2024} or an online stochastic method \citep{he_decision-dependent_2025}.
 
A set of Performative Optimal solutions addresses practical aspects of real-world systems. Recognising that models often operate under constraints, \citep{yan_zero-regret_2023} presented a framework for \textbf{Constraint Optimisation} using a primal-dual stochastic approach, which was later extended to noncooperative multiplayer scenarios in which players react to a deployed model and attempt to improve their position at the expense of others \citep{yan_decentralized_2024}. To handle real-world \textbf{High-Dimensional Models} accurately and efficiently, \citet{chen_practical_2024} proposed focusing on the model itself as the source of change and developing stochastic gradient-based classifiers that scale and converge properly with high-dimensional models.

To solve the ``unknown map" problem without complex gradient estimation, \textbf{Derivative-Free Optimisation (DFO)} methods can bypass the need for exact knowledge of the distribution map by using zeroth-order optimisation \citep{chen_performative_2024}. Although DFO methods are less sensitive to errors in the model specification, they are slow to converge \citep{lin_plug-performative_2024}. To overcome these deficiencies, \citet{lin_plug-performative_2024} proposed a procedure comprising three stages: data collection and exploration, distribution map estimation, and optimal point calculation. Using this procedure enables faster convergence to the optimal point, even in the presence of errors in the model's specification.

Another proposed approach to arrive at a performative optimal solution is to learn the distribution map through a reverse-causal lens \citep{somerstep_learning_2024}, whereby the response of actors in the environment to the deployed model is the cause of the performative distribution shift \citep{bracale_learning_2024}. In this approach, the reaction to the deployed model is framed in terms of the perceived benefits of responding to it, while accounting for the associated response costs \citep{bracale_microfoundation_2024}. Estimating the cost-benefit function of the actors allows inferring the model's distribution map, which, in turn, facilitates the use of fast optimisation algorithms to minimise the performative risk \citep{bracale_microfoundation_2024}.

Following the work of \citet{jin_performative_2024} on performative federated learning, which focused on performative stability and created the \textbf{Performative FedAvg (P-FedAvg)} algorithm, \citep{zheng_profl_2024} presented an algorithm that can arrive at a performative optimal solution. The \textbf{Performative optimal Federated Learning (ProFL)} algorithm can converge to an optimal point while supporting a broader range of performative cases and being more robust to contaminated data than previous algorithms \citep{zheng_profl_2024}.

\subsubsection{Approximate Optimality Solutions}

An approach proposed by \citet{liu_two-timescale_2024} extends beyond finding a stable or optimal solution. It aims to reach a near-stationary point that approximates the performative optimal point without requiring knowledge of the predictive model's loss function. This approach utilises stochastic derivative-free optimisation (DFO) to estimate the gradient of the loss function by evaluating it at sampled points.

Table ~\ref{tab:algo_solutions} summarises and contrasts the algorithmic solutions for performative prediction, organising them by their primary objective (stability vs. optimality), core mechanisms, and key limitations.

{\centering
\fontsize{8pt}{9pt}\selectfont

\begin{longtable}{@{} p{0.10\textwidth} 
                      p{0.15\textwidth} 
                      p{0.265\textwidth} 
                      p{0.28\textwidth} 
                      p{0.15\textwidth} @{}}

\caption{Summary of Algorithmic Solutions for Performative Prediction}
\label{tab:algo_solutions} \\

\toprule
\textbf{Primary Objective} & \textbf{Method/Approach} & \textbf{Key Idea/Mechanism} & \textbf{Main Limitations \& Criticisms} & \textbf{Key Refs.} \\
\midrule
\endfirsthead

\caption[]{Summary of Algorithmic Solutions (continued)} \\
\toprule
\textbf{Primary Objective} & \textbf{Method/Approach} & \textbf{Key Idea/Mechanism} & \textbf{Main Limitations \& Criticisms} & \textbf{Key Refs.} \\
\midrule
\endhead

\midrule
\multicolumn{5}{r}{\textit{Continued on next page}} \\
\endfoot

\bottomrule
\endlastfoot

Stability & Repeated Risk Minimisation (RRM) & Iteratively retrain a model on the data distribution created by the previous model's deployment until it converges to a fixed point. & May fail to converge, or converge to a suboptimal, and potentially unfair stable point. & \citep{perdomo_performative_2020}\citep{hardt_performative_2023}\citep{tsoy_impact_2025}\citep{dong_approximate_2023} \\
\addlinespace
Stability & RRM Variants (e.g., Reg-RRM, R³M) & These methods modify RRM to improve convergence, for instance by adding regularisation to slow the retraining pace or by using a set of potential distributions when the true one is unknown. & They address specific RRM shortcomings but add complexity to the training process. & \citep{mofakhami_performative_2023}\citep{kabra_limitations_2024}\citep{jia_distributionally_2024}\citep{khorsandi_tight_2024} \citep{lu_bilevel_2023}  \\
\addlinespace
Stability & Stochastic Gradient Descent (SGD) and its variants (e.g. Clipped-SGD, Stateful) & Instead of full retraining, it updates the model's parameters using a single gradient step on the loss function. Can be "greedy" (update on every new data point) or "lazy" (update after several). & The choice between greedy and lazy deployment depends on the cost of updating the model versus the severity of the performativity. & \citep{perdomo_performative_2020}\citep{perdomo_performative_2023}\citep{drusvyatskiy_stochastic_2023}\citep{mendler-dunner_stochastic_2020} \citep{li_stochastic_2024}\citep{li_clipped_2024}\citep{cutler_stochastic_2024}\citep{li_state_2022} \citep{brown_performative_2022} \\
\addlinespace
Stability & Multi-Agent / Game-Theoretic Stability & Extends stability concepts to scenarios with multiple competing or cooperating agents, seeking a Nash Equilibrium or a cooperative stable point. & The dynamics can be complex, potentially leading to instability or chaos under certain conditions. & \citep{narang_learning_2022} \citep{narang_multiplayer_2023}\citep{li_multi-agent_2022}\citep{wang_network_2023}\\
\addlinespace
Optimality & Performative Gradient Descent (PerfGD) and its extensions & Directly optimise the performative risk by estimating the gradient of the risk function itself, rather than just seeking a stable point. & Its primary challenge is that the true distribution map induced by the model is unknown and must be estimated, making it sensitive to errors in the model specification. & \citep{izzo_how_2021}\citep{miller_outside_2021}\citep{izzo_how_2022}\citep{izzo_theory_2023} \citep{cyffers_optimal_2024} \citep{zhao_optimizing_2022} \\
\addlinespace
Optimality & Dynamic Environment Optimisation / Regret Minimisation & Finds an optimal point in cases where the data distribution doesn't change immediately but evolves to a stable state after deployment. It uses an iterative stochastic gradient algorithm.& This iterative approach can be slow, as it may require multiple model deployments per update to allow the environment to stabilise before calculating the next step. & \citep{ray_decision-dependent_2022} \\
\addlinespace
Optimality & Online / Regret Minimisation & Frames the problem in an online setting where the goal is to minimise cumulative loss (regret) over time as the model and data distribution co-evolve. & The goal is not to find a single, final "optimal" model but to maintain low regret over time, which is a different objective than standard optimisation. & \citep{he_decision-dependent_2025}\citep{jagadeesan_regret_2022}\citep{park_parameter-free_2024} \\
\addlinespace
Optimality & Constraints / Game-Theoretic Optimisation & Finds an optimal solution in cases where the model's parameters are constrained, or in multi-agent games where players compete & Often computationally expensive, as they require solving complex nested problems at each step. Theoretical convergence guarantees also rely on restrictive assumptions (e.g., strong convexity, monotonicity) that may not hold in practice. & \citep{yan_zero-regret_2023}\citep{yan_decentralized_2024} \\
\addlinespace
Optimality & High-Dimensional Models & Reframes the problem to focus on the model itself, not just its parameters, to develop scalable, gradient-based classifiers for high-dimensional settings. & The main challenge lies in the complexity of analysing the model itself as a function, rather than the more traditional and intuitive analysis of its parameters. & \citep{chen_practical_2024} \\
\addlinespace
Optimality & Zeroth-Order Optimisation & Aims to find an optimal point without needing to know the exact gradient of the loss function; instead, it estimates the gradient by evaluating the loss at sampled points. & It can enable finding optimality without precise knowledge of the distribution map, but is generally much slower to converge than gradient-based methods. & \citep{chen_performative_2024} \\
\addlinespace
Optimality & Reverse Causal / Cost-Benefit Models & Learn the distribution map by inferring the cost-benefit function of strategic agents. This estimated map can then be used in faster, gradient-based optimisation algorithms. & Relies on the ability to accurately model the motivations and strategic behaviour of human agents, which can be difficult to specify correctly. & \citep{lin_plug-performative_2024}\citep{somerstep_learning_2024}\citep{bracale_learning_2024}\citep{bracale_microfoundation_2024} \\
\addlinespace
Stability \& Optimality & Distributionally Robust Optimisation (DRO) Methods & Uses robust optimisation to handle uncertainty in the data distribution. RDRO aims for a fair, stable point, while DRPO aims for an optimal one, especially when the distribution map is misspecified. & Adds the complexity of robust optimisation, requiring the definition of a set of potential distributions, which can be challenging. & \citep{peet-pare_long_2022}\citep{peet-pare_beyond_2022}\citep{xue_distributionally_2024} \\
\addlinespace
Stability \& Optimality & Federated Learning & Adapts performative prediction to a decentralised setting where multiple agents collaboratively train a model. Algorithms like P-FedAvg (stability) and ProFL (optimality) are used. & Inherits the challenges of standard performative prediction while adding the complexities of decentralised training and communication overhead. & \citep{jin_performative_2024}\citep{zheng_profl_2024} \\
\addlinespace
Approximate Optimality & Derivative-Free Optimisation (DFO) & Aims to reach a near-stationary point that approximates the performative optimal point. This is achieved by estimating the gradient of the loss function without needing explicit gradient knowledge of the performative risk. & Can enable finding optimality without precise knowledge of the distribution map, but is generally much slower to converge than gradient-based methods. & \citep{liu_two-timescale_2024} \\

\end{longtable}}

\subsection{Non-Algorithmic Solutions}

In contrast to the algorithmic solution detailed in the previous section, a complementary body of work addresses the challenges of performativity through higher-level, non-algorithmic solutions. These solutions focus less on optimising a specific loss function and more on a model's conceptual framing, real-world monitoring, and alignment with broader goals. This section reviews these solutions, beginning with conceptual re-framing (Section 6.2.1), followed by detection and monitoring (Section 6.2.2), and concluding with systems and design interventions (Section 6.2.3). To synthesise these solutions, Table \ref{tab:non_algorithmic_solutions} provides a comparative summary at the end of this section.

\subsubsection{Conceptual Re-framing}

The research in this area proposes that performativity can often be addressed by re-framing the problem, such as through causal reasoning or the development of new evaluation frameworks. One line of work suggests that performative shifts might be avoided altogether under certain conditions. For instance, \citep{kulynych_causal_2022} argued that using only \textbf{causal features} for predictions might lead to stability without retraining, provided the model's deployment only affects the predictive features and not the target variable itself.

However, performativity often does affect the target variable, potentially leading to bias and unfairness. Recognising this, other conceptual approaches aim to correct these issues. \citet{boeken_evaluating_2024} conceptualised model deployment as a \textbf{causal domain shift}, offering methodologies to assess and potentially correct for the resulting performative bias, though acknowledging that this may require using randomised testing, which may not be ethical or recommended in some cases \citep{boeken_evaluating_2024}. Similarly, \citet{mishler_fair_2022} noted that models that were fair during training can become unfair when deployed due to performativity and suggested targeting \textbf{counterfactual outcomes} rather than observable results during training as a potential solution. Going a step further, \citet{wyllie_fairness_2024} proposed using \textbf{Algorithmic Reparation (AR)}, leveraging performativity itself via specialised sampling algorithm, \textbf{STratified Sampling Algorithmic Reparation (STAR)}, to actively promote better representation for marginalised groups

Beyond direct interventions, some research focuses on robust evaluation within the performative setting. \citet{li_statistical_2025} established a framework for statistical inference under performativity, including valid confidence levels and hypothesis testing, by using 
\textbf{Prediction-Powered Inference (PPI)} \citep{zrnic_prediction_2023}, which combines a small set of ground-truth labels with a larger set of model predictions to improve estimation accuracy. This provides tools for quantifying uncertainty and testing hypotheses while accounting for the effects of performativity. Complimenting this, \citet{cheng_causal_2024} proposed a framework to evaluate the impact of performativity on digital platforms, avoiding randomised tests by analysing user interactions and measuring changes in consumption behaviour over time.
 
Finally, \citet{makowski_performative_2025} reframed the performativity problem at the feature level, suggesting the use of neural networks to create \textbf{drift-resistant feature representations} that map performatively shifted data back towards its original distribution.

\subsubsection{Detection And Monitoring}

Instead of reframing the problem, the solutions in this category focus on detecting and monitoring the effects of performativity in deployed systems.

A key challenge is obtaining unbiased data for evaluation once a model has begun to actively influence outcomes. One proposed solution involves using \textbf{hold-out sets}, where a portion of the population is intentionally excluded from the model's influence (e.g. receiving standard care regardless of the model's prediction), allowing their outcomes to be used for unbiased retraining \citep{chislett_ethical_2024}. While effective, this approach raises significant ethical concerns, especially in high-stakes domains, that need to be weighed against its potential benefits \citep{chislett_ethical_2024}.

Given the difficulties in implementing hold-out sets, for instance, in our examples of the readmission model and the prognostic mortality model, other approaches focus on monitoring using the already available, performatively influenced data. \citet{feng_monitoring_2024} presented a framework for monitoring the impact of predictive models deployed in healthcare settings, focusing on conditional metrics rather than overall model performance. This framework has been incorporated into a broader framework to monitor performativity using causal reasoning \citep{feng_designing_2024}.

Another technique attempts to anticipate the performative effects directly. The \textbf{Predicting From Predictions} method uses the model's own predictions as an input feature, alongside its other inputs, aiming to foresee the eventual performative outcome, assuming that the model's causal effect is identifiable \citep{mendler-dunner_anticipating_2022}.

Finally, it is essential to distinguish between performative effects and other changes. \textbf{CheckerBoard Performative Drift Detection (CB-PDD)} offers a method specifically designed to detect drift in data streams and identify whether that drift was caused by the model's performativity or other external factors \citep{gower-winter_identifying_2025}.

\subsubsection{Systems And Design Interventions}

This final category moves beyond monitoring to advocate for proactive, human-centric design choices that align a model's predictive function with its ultimate real-world objective.

Predictive models can cause harm through self-fulfilling or self-negating prophecies, even when they were well-trained and validated to achieve a positive outcome \citep{amsterdam_when_2025}. To prevent this, \citet{amsterdam_when_2025} called for a shift towards \textbf{Casual Alignment}, particularly in high-stakes areas like healthcare. This involves designing and validating models not only for predictive accuracy, but explicitly for their ability to improve the desired outcomes (e.g. patient health) by incorporating causal reasoning throughout the development process. In the case of the prognostic mortality model, instead of designing a model to predict death passively, this approach advocates for developing a model to actively achieve the clinical goal.

As this section has shown, non-algorithmic solutions offer a different set of approaches, focusing on the framing, monitoring, and designing of predictive systems. To provide a consolidated overview of these approaches, Table \ref{tab:non_algorithmic_solutions} summarises their core ideas and mechanisms, as well as their limitations.

{ 
\fontsize{8pt}{9pt}\selectfont

\centering
\begin{longtable}{@{} p{0.10\textwidth} 
                      p{0.15\textwidth} 
                      p{0.265\textwidth} 
                      p{0.28\textwidth} 
                      p{0.11\textwidth} @{}}

\caption{Summary of Conceptual, Monitoring, and Design Solutions}
\label{tab:non_algorithmic_solutions} \\

\toprule
\textbf{Category} & \textbf{Method/Approach} & \textbf{Key Idea / Mechanism} & \textbf{Main Limitations \& Criticisms} & \textbf{Key Refs.} \\
\midrule
\endfirsthead
\caption[]{Summary of Conceptual, Monitoring, and Design Solutions (continued)} \\
\toprule
\textbf{Category} & \textbf{Method/Approach} & \textbf{Key Idea / Mechanism} & \textbf{Main Limitations \& Criticisms} & \textbf{Key Refs.} \\
\midrule
\endhead

\midrule
\multicolumn{5}{r}{\textit{Continued on next page}} \\
\endfoot

\bottomrule
\endlastfoot

Conceptual Re-framing & Causal Features & Proposes that using only causal features for prediction can, under certain conditions, lead to performative stability without needing to retrain the model after deployment. & This approach assumes that the model's deployment only affects its predictive features and not the target variable it is trying to predict. & \citep{kulynych_causal_2022} \\
\addlinespace
Conceptual Re-framing & Causal Frameworks (Domain Shift / Counterfactual)  & Conceptualises performativity as a causal domain shift or targets counterfactual outcomes (what would have happened) to assess and correct for performative bias and unfairness. & These approaches can be challenging to implement, as they may require randomised testing, which may be unethical or undesirable or the estimation of unobserved counterfactual outcomes. & \citep{mishler_fair_2022}\citep{boeken_evaluating_2024} \\
\addlinespace
Conceptual Re-framing & Algorithmic Reparation (AR) & Leverages the mechanism of performativity to intentionally create positive social outcomes by using special sampling methods (like STAR) to ensure better representation for marginalised groups. & Requires a non-technical definition of "equity", risking misinterpretation as a mere technical fix. In practice, it may cause a trade-off with the accuracy of the predictive model, and the sampling methods used may increase, over time, the effects of mislabelling and bias & \citep{wyllie_fairness_2024} \\
\addlinespace
Conceptual Re-framing & Prediction-Powered Inference (PPI) & Establishes a formal framework for constructing valid confidence intervals and conducting hypothesis testing in performative settings by combining a small set of ground-truth labels with a larger set of model predictions. & Relies on strong theoretical assumptions, and can be computationally intensive, and is currently limited mainly to data-scarce scenarios.  & \citep{li_statistical_2025,zrnic_prediction_2023} \\
\addlinespace
Conceptual Re-framing & Observational Evaluation Framework & Proposes a framework to evaluate performativity on digital platforms by observing user interactions over time, avoiding the need for randomised tests. & Relies on observational data, which may be subject to confounding variables, making causal claims difficult. & \citep{cheng_causal_2024} \\
\addlinespace
Conceptual Re-framing & Drift-Resistant Feature Representations & Uses neural networks to create feature representations that are robust to performative data drift, mapping the induced distribution back to the original one. & Computationally costly and requires a sufficient amount of clean data. Due to the use of neural networks, it is non-interpretable, and its effectiveness reduces if the direction of the performative drift changes, as it is trained to learn only a single mapping. & \citep{makowski_performative_2025} \\
\addlinespace
Detection \& Monitoring & Hold-out Sets & A portion of the population is intentionally excluded from the model's influence (e.g., they do not receive a specific treatment based on the model's prediction), and their outcomes are used for unbiased model retraining. & While this can mitigate risks associated with performativity, it raises significant ethical questions and considerations, especially in high-stakes domains like medicine. & \citep{chislett_ethical_2024} \\
\addlinespace
Detection \& Monitoring & Conditional Metrics \& Causal Reasoning & A framework for monitoring deployed models by focusing on performance metrics for specific subgroups (conditional metrics) rather than just overall model performance, using causal reasoning to navigate performativity. & Relies on strong causal assumptions that the model might violate, and implementation can be complex, requiring pre-monitor studies. & \citep{feng_monitoring_2024,feng_designing_2024} \\
\addlinespace
Detection \& Monitoring & Predicting From Predictions & Uses the model's own predictions as an input feature, alongside its other inputs, to anticipate the performative outcome. & Relies on the premise that the causal effect of the model is identifiable & \citep{mendler-dunner_anticipating_2022} \\
\addlinespace
Detection \& Monitoring & Performative Drift Detection (CB-PDD) & A method designed to detect if a change in a data stream's distribution was caused by the model's performativity or by other external factors. & Implementation will require deliberately misclassifying a portion of incoming instances, which may be infeasible or unethical in real-world settings. & \citep{gower-winter_identifying_2025} \\
\addlinespace
Systems \& Design Interventions & Causal Alignment & Calls for a shift in model design for high-stakes scenarios to focus on aligning the models with the ultimate objective (e.g., improving patient outcomes) through causal reasoning, rather than just predictive accuracy. & Implementation requires a fundamental redesign of machine learning practices. It introduces an ethical dilemma by introducing a definition of a "desired" outcome, and the practical evaluation may require expensive controlled trials or strong causal assumptions.  & \citep{amsterdam_when_2025} \\
\bottomrule
\end{longtable}
}

%% file: Chapters/PP-Extensions.tex
\section{Extensions To Performative Predictions}

While the foundational research focused mainly on the deployment of a single supervised learning model \citep{narang_learning_2022,li_multi-agent_2022,wang_multi-agent_2025}, performativity also arises in more complex scenarios. This section surveys important extensions to the foundational research, highlighting unique challenges in multi-agent systems (including human-ML collaboration) and various machine learning paradigms. 

\subsection{Multi-agent Performative Predictions}

A significant body of research explores the \textbf{multi-agent} context, where multiple predictive models interact with the same population \citep{narang_learning_2022}, introducing new complexities beyond the single-agent setting.

In \textbf{competitive scenarios}, such as universities using separate admissions models \citep{narang_learning_2022} or financial institutions predicting the same market outcome \citep{piliouras_multi-agent_2023}, the dynamics can become unstable \citep{piliouras_multi-agent_2023}. Even small changes in the behaviour of the agents can cause significant changes in the data distributions \citep{wang_multi-agent_2025}. Achieving \textbf{Performative Stability}, often defined as a Nash Equilibrium, can be done using adaptation of single-agent solutions. Methods based on repeated retraining or stochastic gradients (as discussed in Section 6.1.1) have been developed for this purpose \citep{narang_learning_2022,narang_multiplayer_2023}. Another challenge unique to competitive settings is dishonest reporting, which can distort the shared environment \citep{hudson_joint_2025}. A potential solution is to use a zero-sum competition with scoring rules that incentivise honest reporting by all participating agents \citep{hudson_joint_2025}. Finally, a recent work by \citep{le_cadre_learning_2025} focused on achieving stability in situations where competing react to the results of a deployed model while keeping some of their information private.

Conversely, multi-agent scenarios can also be \textbf{cooperative}, such as when healthcare providers collaborate to develop a predictive model using their separate datasets to benefit their respective populations, while potentially achieving better generalisation and robustness \citep{li_multi-agent_2022}. Solutions here often involve decentralised algorithms (discussed in Section 6.1.1 \citep{li_multi-agent_2022}) or specialised frameworks, such as Federated Learning (discussed in Section 6.1.2 \citep{jin_performative_2024,zheng_profl_2024}), to achieve stable or optimal joint models. Network effects, where agents learn from each other's deployments, have been studied by \citep{wang_network_2023}, who demonstrated that both a performative stable solution and a Nash Equilibrium can be achieved using a distributed stochastic gradient descent method.

Going beyond model-to-model interactions, \citep{gois_performative_2024} focused on connected predicted outcomes and the risk of suboptimal collective outcomes, even when the predictive models are accurate. For instance, the case of different individuals reacting to predictive pandemic spread models. The paper proposed a method to understand the population's response to the deployed model, thereby directing the environment towards a more positive social outcome.

Finally, a related line of inquiry explores \textbf{human-ML collaboration}, which can be understood as a specific type of multi-agent dynamics where the model's predictions influence human users, and the model, in turn, learns from the humans' feedback. \citep{suhr_dynamic_2024} modelled this as a dynamic process where predictive models learn from human input that is itself influenced by the deployed models, and showed that convergence to a stable point is possible, albeit some may be suboptimal.

\subsection{Other Machine Learning Methods}

The core concepts of performativity have also been adapted beyond the supervised learning paradigms.

\begin{itemize}
    \item \textbf{Time-Series Forecasting}: Here, performativity presents a unique challenge as the predictions directly influence future observations in the sequence. \citep{zhao_performative_2023} coined the term \textbf{Performative Time-Series Forecasting (PeTS)} and developed specific methods like Feature Performative-Shifting (FPS) that uses delayed responses to predict changes in data distribution and the ensuing predicted outcomes.
    \item \textbf{Reinforcement Learning (RL)}: In \textbf{Performative Reinforcement Learning (PRL)}, the environment itself changes in response to the RL agent's deployed policy \citep{mandal_performative_2023}. Achieving stable policies can be achieved by adapting repeated retraining methods \citep{mandal_performative_2023}. Subsequent works extended and generalised PRL to larger-scale, realistic use-cases \citep{mandal_performative_2024}, and to environments that adapt gradually to the deployed policy \citep{rank_performative_2024}.
    \item \textbf{Deep Learning}: Extension of performative predictions to deep learning models was introduced by \citep{demirel_adjusting_2024}, who argued that the standard methodologies to account for performativity would not work in the case of deep learning models due to the amount of data necessary for retraining the models, and the use of a direct features-labels connection that does not exist in deep learning models. To adjust for performativity that causes a change in the split between classes, \citep{demirel_adjusting_2024} suggested adding an adaptation module to the structure of the pre-trained model, allowing it to adapt its predictions to the performativity.
\end{itemize}

%% file: Chapters/Performative-Matrix.tex
\section{Performative Strength vs. Impact Matrix}

The preceding sections have mapped the landscape of performative predictions, detailing the mechanisms through which it arises (Section 4), the various risks it creates (Section 5), and the technical and conceptual solutions proposed to manage it (Section 6). However, a significant challenge remains for practitioners: how to reason about a specific, real-world use case and select an appropriate strategy to manage potential performativity.
To bridge this gap, we introduce the \textbf{Performative Strength vs. Impact Matrix} - a novel conceptual framework for assessing the nature and severity of performativity in real-world scenarios. The matrix provides a structured approach for evaluating a model's potential to influence its environment and the consequences of that influence, thereby guiding decisions on governance, monitoring, and mitigation if required.
The matrix positions use cases along two dimensions: \textbf{Performativity Strength} and \textbf{Performativity Impact}. Performativity Strength represents the extent to which the deployment of a model causes a change in the data distribution it later trains or evaluates on. Performativity Impact represents the expected magnitude and severity, either positive or negative, of the outcomes attributed to performativity. Together, these dimensions help to assess and evaluate the consequences of deploying predictive models.

We assign each of these dimensions one of three values: low, medium, and high, to create a nine-cell matrix that is rich enough to capture the diversity of predictive models' use cases, yet simple enough to be adapted as a practical decision-making tool.

\begin{figure}[htbp]
  \centering
  \includesvg[width=0.4\linewidth]{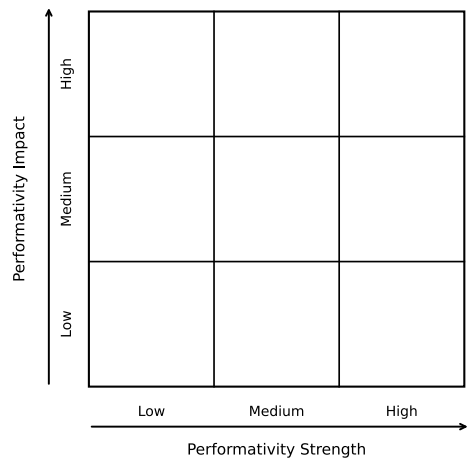}
  \caption{Performativity Strength / Impact Matrix}
  \label{fig:performative-matrix}
\end{figure}

\textbf{Definitions:}

We intuitively define the levels for performative strength as follows:
\begin{itemize}
    \item \textbf{Low performative strength} - predictions have little influence on the environment. The outcomes remain largely unaffected by the prediction, and any distribution shift is negligible.
    \item \textbf{Medium performative strength} - predictions shape the environment in noticeable ways; however, the effects are partial or restricted. Models' deployment may induce a moderate distribution shift; yet, the overall system remains mostly stable.
    \item \textbf{High performative strength} - the predictions strongly drive behavioural or systematic changes. Outcomes become highly entwined with the act of prediction, often creating feedback loops and significant distribution shifts.
\end{itemize}

Similarly, we define the levels for performative impact:

\begin{itemize}
    \item \textbf{Low performative impact} - even if performativity occurs, its consequences are minor, limited to a small number of features or users, or short-term dynamics. The system's performance and risk remain essentially unchanged. 
    \item \textbf{Medium performative impact} - consequences are more widespread, affecting more features, users, or processes. Changes to performance or risks are more evident, but not critical.
    \item \textbf{High performative impact} - consequences are broad and systematic. Distribution shifts or behavioural changes cascade through the environment, raising significant risks or creating new opportunities.
\end{itemize}

An important distinction is to be made here between \textbf{performative impact} and \textbf{societal impact}. The \textbf{Impact} axis in our matrix refers only to the consequences arising from a model's performative nature, the effects caused by the model changing the data-generating process. It does not refer to the general, real-world impact of the prediction itself.
Consider a weather forecast model predicting a major hurricane. The societal impact of the prediction is immense. An accurate forecast can save lives through proper preparation and evacuations. However, the performative impact is negligible. The forecast does not change the path of the hurricane or its intensity, nor does it change the underlying meteorological data-generating process. Therefore, when assessing a model using the matrix, we focus only on the consequences that stem from performativity, and not the broader importance of the prediction.

To make the conceptual framework of the Performative Strength vs Impact Matrix more concrete, Table \ref{tab:matrix_examples} provides a real-world example for each of the nine cells, alongside a \textbf{potential strategy} derived from the solutions surveyed in this work. It is important to note that placing an instance into a specific cell is a subjective assessment. The boundaries between "Low", "Medium", and "High" are not rigid, and a given use case could be argued to fall in an adjacent cell depending on its deployment context. Further more, the strategies listed are potential recommendations; the actual solution for any specific real-world problem will heavily depend on its unique context and constrains..The following table is intended to be illustrative rather than definitive or rigid in classification.

{ 
\centering
\fontsize{8pt}{9pt}\selectfont

\begin{longtable}{@{} p{0.08\textwidth} 
                      p{0.08\textwidth} 
                      p{0.15\textwidth} 
                      p{0.20\textwidth} 
                      p{0.20\textwidth}
                      p{0.20\textwidth} @{}}

\caption{Performative Strength vs. Impact Matrix Examples}
\label{tab:matrix_examples} \\

\toprule
\textbf{Performative Strength} & \textbf{Performative Impact} & \textbf{Real-World Example} & \textbf{Strength Rationale} & \textbf{Impact Rationale} & \textbf{Potential Strategy} \\
\midrule
\endfirsthead

\caption[]{Performative Strength vs. Impact Matrix Examples (continued)} \\
\toprule
\textbf{Performative Strength} & \textbf{Performative Impact} & \textbf{Real-World Example} & \textbf{Strength Rationale} & \textbf{Impact Rationale} & \textbf{Potential Strategy} \\
\midrule
\endhead

\midrule
\multicolumn{6}{r}{\textit{Continued on next page}} \\
\endfoot

\bottomrule
\endlastfoot

Low & Low & Earthquake Aftershock Prediction & The model's prediction does not influence the underlying geological process. & The performative impact is negligible as the prediction doesn't change the event's outcome. &\textbf{Standard Drift Detection.} No causal effect means that performative algorithmic solutions are not needed. Monitor for external data shifts.  \\
\addlinespace
Low & Medium & Retail Inventory Demand Forecasting & Demand is primarily driven by external factors rather than by the stocking decision itself & Inaccurate predictions lead to moderate financial losses through spoilage (overstocking) or lost revenue (stockouts) &\textbf{Automated Retraining \& Monitoring.} Since the feedback loop is negligible, standard automated retraining is safe. Financial stakes justify tighter monitoring thresholds for external shifts.\\
\addlinespace
Low & High & One-Time Market Exploit Model & It's a single-use prediction. Once the exploit is used, no sustained feedback loop is created for retraining. & The action taken based on the prediction (the exploit) permanently and significantly alters market rules and outcomes. &\textbf{Manual Redesign \& Causal Reasoning (Section 6.2.1).} Avoid automated retraining, and instead model the new market structure.\\
\addlinespace
Medium & Low & Personal Music Recommendation & The model's recommendations can influence the user's listening habits, and the resulting behavioural data is fed back into the model. & The consequences are minor and personal, affecting only an individual's musical taste or entertainment preferences. &\textbf{Repeated Risk Minimisation (Section 6.1.1).} Automated RRM effectively allows adaptation to shifting tastes, converging to a stable profile.\\
\addlinespace
Medium & Medium & E-Commerce Recommendation Engine & The model's recommendations noticeably steer user purchases, creating a feedback loop that alters sales data and product rankings. & The consequences are widespread enough to have real financial effects on third-party sellers and influence the marketplace. &\textbf{Performative Optimisation (Section 6.1.2).} Use optimisation algorithms such as \textbf{PerfGD} to steer the distribution towards a global optimum (e.g., long-term user value). \\
\addlinespace
Medium & High & Pre-trial Bail/Detention Model & The model predicts a "risk score" that a judge consults to decide on bail. The mediating effect of the judge's human-made decision reduces the model's performative strength. & The consequences of the decision are severe, including the loss of liberty. The model also has the potential to entrench biases and affect marginalised communities. &\textbf{Algorithmic Reparation (Section 6.2.1).} Avoid RRM due to bias risk. Use \textbf{STAR} to sample underrepresented groups and prevent bias amplification and entrenchment. \\
\addlinespace
High & Low & In-Game Non-Player Character Behaviour & The AI characters' behaviour is driven by the player's actions, creating a real-time feedback loop that defines the gameplay. & The consequences are entirely contained within a low-stakes, virtual environment with no real-world harm. &\textbf{Performative RL (Section 7.2).} Use \textbf{PRL} to learn policies that adapt to player tactics in real-time.\\
\addlinespace
High & Medium & Dynamic Surge Pricing (e.g., Uber) & The model's prediction directly determines a new price, which directly changes users' and drivers' behaviours in a strong, fast, feedback loop & The consequences are financial and often widespread, potentially affecting a large number of users and drivers. However, the impact is typically not life-changing or systemic. &\textbf{Game-Theoretic Stability Section (6.1.1).} There is a risk of price oscillation. Use \textbf{Multi-Agent Stability} to find Nash Equilibrium.\\
\addlinespace
High & High & Credit Scoring Model & The model's prediction directly causes the outcome it seeks to predict, creating a powerful, self-fulfilling prophecy. & The consequences are severe, systemic, and potentially leading to financial exclusion and entrenching inequality. &\textbf{Causal Alignment (Section 6.2.3).} Use Causal Alignment to design the model for adequate financial health.\\
\end{longtable}
} 

Table 3 illustrates that, although each use case is unique, the potential strategies tend to cluster into three distinct zones. To resolve potential overlaps between the zones, we define them in a hierarchical structure, where the severity of the impact dictates the strategy first, followed by the performative strength. These three zones are visually summarised in Figure \ref{fig:matrix_zones} and further detailed below.
\input{figures/matrix_zones.tex}

\textbf{Zone 1: The Observation Zone - } This zone includes the \textbf{Low Strength / Low-Medium Impact} cell, where the model has negligible causal influence on the data-generating process and the societal stakes are low to moderate. In the absence of a strong feedback loop, performative-specific algorithms add unnecessary complexity. The recommended strategy for this zone is robust monitoring to detect any external data drifts, ensuring the model remains calibrated to the external world.

\textbf{Zone 2: The Algorithmic Management Zone - } This zone covers the \textbf{Medium-High Strength / Low Impact} cells, where the effects of the feedback loop are strong (the model actively shapes the data), but the societal consequences are contained. Given the risk's manageability, practitioners can safely leverage automation with stability-seeking algorithms, such as RRM, or optimisation-seeking algorithms, such as PerfGD.

\textbf{Zone 3: The Socio-Technical Governance Zone - } This zone includes all \textbf{High Impact} scenarios, regardless of their performative strength (credit scoring, bail decisions, or market exploits). Here, the costs of error are too high to rely only on automated algorithms. In this zone, governance takes precedence over automation, and practitioners should avoid "black-box" retraining and instead redesign and realign models to avoid social harm.

%% file: figures/matrix_zones.tex
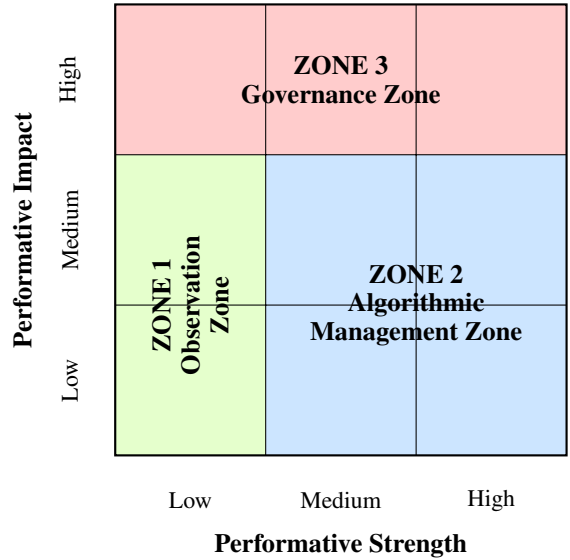
\begin{figure}[ht]
    \centering
    \begin{tikzpicture}[scale=2]
        \definecolor{ZoneGov}{RGB}{255, 204, 204}   
        \definecolor{ZoneAlgo}{RGB}{204, 229, 255}  
        \definecolor{ZoneObs}{RGB}{229, 255, 204}   

        \fill[ZoneGov] (0,2) rectangle (3,3);
        
        \fill[ZoneAlgo] (1,0) rectangle (3,2);
        
        \fill[ZoneObs] (0,0) rectangle (1,2);

        \draw[thick] (0,0) rectangle (3,3); 
        \draw[step=1.0, black, thin] (0,0) grid (3,3); 

        \node at (1.5, -0.6) {\textbf{Performative Strength}};
        \node at (0.5, -0.3) {\small Low};
        \node at (1.5, -0.3) {\small Medium};
        \node at (2.5, -0.3) {\small High};

        \node[rotate=90] at (-0.6, 1.5) {\textbf{Performative Impact}};
        \node[rotate=90] at (-0.3, 0.5) {\small Low};
        \node[rotate=90] at (-0.3, 1.5) {\small Medium};
        \node[rotate=90] at (-0.3, 2.5) {\small High};

        \node[align=center] at (1.5, 2.5) {\textbf{ZONE 3} \\ \textbf{Governance Zone}};
        
        \node[align=center] at (2.0, 1.0) {\textbf{ZONE 2} \\ \textbf{Algorithmic} \\ \textbf{Management Zone}};

        \node[align=center, rotate=90] at (0.5, 1.0) {\textbf{ZONE 1} \\ \textbf{Observation} \\ \textbf{Zone}};

    \end{tikzpicture}
    \caption{Performative Strength vs. Impact Matrix Zones}
    \label{fig:matrix_zones}
\end{figure}

%% file: Chapters/Discussion.tex
\section{Discussion}

This Synthesis of Knowledge (SoK) has detailed the evolving landscape of performative predictions, moving from foundational concepts to mechanisms, risk, and solution strategies. The discussion below synthesises these findings, clarifies the primary contributions of the work, and outlines key limitations and avenues for future research.

\subsection{What the SoK clarifies}

This work systemises the field of performative predictions by organising it around three core axes: the \textbf{mechanisms} of performativity (Section 4), the \textbf{risks} it creates (Section 5), and the diverse \textbf{solutions} proposed to manage it (Section 6). 

Revisiting the core research questions defined in Section 3.1, our work clarifies:

\begin{itemize}
    \item  \textbf{RQ1 - Mechanisms:} Performativity manifests primarily through \textbf{feedback loops} that fundamentally change the underlying data distribution, creating internal \textbf{data shifts} that violate the conventional machine learning models' assumptions.
    \item \textbf{RQ2 - Risks:} The risks associated with performativity are inherently socio-technical. They range from \textbf{performance failures} (e.g., misestimation of risk, oscillation) to severe \textbf{societal harms}, including the creation of harmful self-fulfilling prophecies and entrenchment of bias.
    \item \textbf{RQ3 - Strategies:} Mitigation strategies fall into two primary categories: \textbf{Algorithmic Solutions}, which strive to manage performativity mathematically (seeking either Stability or Optimality), and \textbf{Non-Algorithmic Solutions}, which focus on governance, monitoring, and causal alignment.
\end{itemize}

Building on these direct answers to the research questions, our work reveals several broader key insights:

\begin{itemize}
    \item \textbf{A fundamental tension in objectives}: A recurring theme is the tension between \textbf{predictive accuracy} and \textbf{outcome steering}. Many models are deployed not only to predict the future passively but to actively change it in the direction of a desirable objective, such as preventing a medical condition.
    \item \textbf{Predictive stability vs optimality}: Most of the algorithmic solutions proposed in the literature can be primarily divided into two types: those seeking \textbf{performative stability}, i.e. a point of equilibrium, and those aiming for \textbf{performative optimality}, i.e. the best possible solution. The review shows that a stable point is not necessarily optimal,  but can instead represent a suboptimal equilibrium. This distinction is important for practitioners, as simply retraining a model until it converges to a stable point may fail to achieve the intended objective of the predictive model.
    \item \textbf{From theory to practice with the Strength vs. Impact Matrix}: While the literature is rich with algorithmic solutions, there is less guidance for practitioners on how to reason about the performativity in real-world use cases. The \textbf{Performative Strength vs. Impact Matrix} we presented in Section 8 bridges this gap. By assessing how strong a model influences its environment (\textbf{strength}) and the severity of its consequences (\textbf{impact}), the matrix provides a framework for risk assessment and required actions. We then connect this framework to concrete real-world examples (Table \ref{tab:matrix_examples}) and map the solution landscape into three distinct zones (Figure \ref{fig:matrix_zones}). This integration transforms the matrix into a practical decision-support tool, empowering practitioners to shift from abstract diagnosis to selecting appropriate solutions.
\end{itemize}

\subsection{Limitations and Future Research Directions}
While this work provides an extensive review of performative predictions, this field is continually advancing. We identify several limitations of this work and directions for future research.

\textbf{Limitation of this review}

\begin{itemize}
    \item \textbf{Temporal scope}: Our search was restricted to publications between 2019 and 2025 to capture the most recent developments since the term was formally introduced. Foundational works on feedback loops or strategic behaviour in other fields that predate this period may offer additional understandings.
    \item \textbf{Keyword specificity}: Our search focused on the explicit term "performative prediction". Related concepts have their own bodies of literature that were only partially covered if they did not use the specific search terms.
\end{itemize}

\textbf{Future research direction}

\begin{itemize}
    \item \textbf{Empirical validation and case studies}: Many of the proposed algorithmic solutions presented in the literature have been demonstrated on theoretical models or synthetic data. There is a need for more empirical studies that apply and compare these solutions in real-world cases to understand their practical performance, scalability, and robustness.
    \item \textbf{Governance and non-algorithmic solutions}: The literature in the field is heavily focused on algorithmic solutions. More research is needed on the role of governance, regulations, and human-in-the-loop systems as additional strategies to manage performativity.
    \item \textbf{Long-term and systematic effects}: Most of the current focus in the field is on near-term solutions. Further research is needed on the long-term impacts of performativity. For example, how do feedback loops in predictive models used for hiring, deployed and used over time, affect society through potential entrenchment of inequality? 
    \item \textbf{Developing practical tools for the Strength vs. Impact Matrix}: While the Strength vs. Impact Matrix serves as a conceptual guide, further work is needed to develop practical diagnosis tools to help practitioners identify which "Zone" their use cases occupy. In addition, further work could explore the transition points between cells or zones, identifying indicators for when a system shifts, for instance, from a monitoring-only state (Zone 1) to a state requiring algorithmic management (Zone 2).
    
\end{itemize}

%% file: Chapters/Conclusion.tex
\section{Conclusion}

 Performative prediction represents a shift in how predictive models are viewed, from passively making observations and predictions to actively shaping their environment. This SoK provides a comprehensive overview of this emerging field, including the mechanisms of performativity, risks, and the array of solutions developed to manage its effects.
 The Performative Strength vs. Impact Matrix introduced in this work serves as a bridge between theory and practice, providing practitioners with a framework to consider the potential effects of their predictive models. By evaluating a model's potential to change its environment and the severity of those changes, stakeholders can make more informed decisions about governance and mitigation.
 As machine learning models become increasingly integrated into society, it is essential to understand their potential performative effects. Moving forward, the field needs to continue developing not only algorithmic solutions but also practical governance frameworks and an understanding of the possible long-term effects of predictive models.

%% file: Chapters/List-of-Acronyms.tex
\section*{List of Acronyms}
\addcontentsline{toc}{section}{List of Acronyms}

\begin{longtable}{@{}ll@{}}
\textbf{Acronym} & \textbf{Definition} \\
\midrule
\endhead

AI & Artificial Intelligence \\
AR & Algorithmic Reparation \\
Bi-RRM & Bi-level Repeated Risk Minimisation \\
Bi-SGD & Bi-level Stochastic Gradient Descent \\
BPS & Bi-level Performative Stability \\
CB-PDD & CheckerBoard Performative Drift Detection \\
DFO & Derivative-Free Optimisation \\
DRO & Distributionally Robust Optimisation \\
DRPO & Distributionally Robust Performative Optimisation \\
DSGD-GD & Decentralised Stochastic Gradient Descent (Greedy Deployment) \\
FPS & Feature Performative-Shifting \\
ML & Machine Learning \\
PD & Performative Drift \\
PeTS & Performative Time-Series Forecasting \\
PerfGD & Performative Gradient Descent \\
P-FedAvg & Performative FedAvg \\
PO & Performative Optimality \\
PPI & Prediction-Powered Inference \\
PR & Performative Risk \\
PRL & Performative Reinforcement Learning \\
ProFL & Performative Optimal Federated Learning \\
R\textsuperscript{3}M & Repeated Robust Risk Minimisation \\
RDRO & Repeated Distributed Robust Optimisation \\
Reg-RRM & Regularised Repeated Risk Minimisation \\
RL & Reinforcement Learning \\
RPPerfGD & Reparametrisation-based Performative Gradient \\
RRM & Repeated Risk Minimisation \\
SFB & Stochastic Forward-Backward \\
SGD & Stochastic Gradient Descent \\
SGD-DG & Stochastic Gradient Descent (Greedy Deployment) \\
SoK & Systematisation of Knowledge \\
STAR & STratified Sampling Algorithmic Reparation \\

\end{longtable}